\RequirePackage{amsmath}
\RequirePackage{fix-cm}
\documentclass[]{article}[14-12-2018]
\usepackage{scrextend}
\usepackage[utf8]{inputenc}
\usepackage{amsfonts}
\usepackage{mathtools}
\usepackage{tabularx}
\usepackage{graphicx}
\usepackage{subcaption}
\usepackage{wrapfig}
\usepackage{hyperref}
\usepackage{hyphenat}
\usepackage{listings}
\usepackage{color}
\usepackage{xcolor}
\usepackage[toc,page]{appendix}
\usepackage{multirow}
\usepackage{booktabs}
\usepackage[round]{natbib}
\begin{document}

\title{Deep Learning for Classification Tasks on Geospatial Vector Polygons}
\author{R.H. van 't Veer\footnote{Vrije Universiteit Amsterdam, Kadaster, Geodan: r.h.vant.veer@vu.nl, orcid: \url{https://orcid.org/0000-0003-0520-6684}} \and P. Bloem\footnote{Vrije Universiteit Amsterdam, orcid: \url{https://orcid.org/0000-0002-0189-5817}} \and E.J.A. Folmer\footnote{University of Twente, Kadaster, orcid: \url{https://orcid.org/0000-0002-7845-1763}}}
%\institute{R.H. van 't Veer 
%			\at Vrije Universiteit Amsterdam, De Boelelaan 1105, 1081 HV Amsterdam
%            \at Kadaster, Hofstraat 110, 7311 KZ Apeldoorn
%            \at Geodan, President Kennedylaan 1, 1079 MB Amsterdam
%              \email{r.h.vant.veer@vu.nl} \\
%              orcid: \url{https://orcid.org/0000-0003-0520-6684} %  \\
%           P. Bloem \at
%              Vrije Universiteit Amsterdam, De Boelelaan 1105, 1081 HV Amsterdam \\
%              orcid: \url{https://orcid.org/0000-0002-0189-5817}
%		   E.J.A. Folmer 
%           \at University of Twente, Drienerlolaan 5, 7522 NB Enschede
%           \at Kadaster, Hofstraat 110, 7311 KZ Apeldoorn \\
%            orcid: \url{https://orcid.org/0000-0002-7845-1763}
%}
\maketitle

\begin{abstract}
In this paper, we evaluate the accuracy of deep learning approaches on geospatial vector geometry classification tasks. The purpose of this evaluation is to investigate the ability of deep learning models to learn from geometry coordinates directly. Previous machine learning research applied to geospatial polygon data did not use geometries directly, but derived properties thereof. These are produced by way of extracting geometry properties such as Fourier descriptors. Instead, our introduced deep neural net architectures are able to learn on sequences of coordinates mapped directly from polygons. In three classification tasks we show that the deep learning architectures are competitive with common learning algorithms that require extracted features. 
\end{abstract}

\section*{Acknowledgements}
This work was supported by the Dutch National Cadastre (Kadaster) and the Amsterdam Academic Alliance Data Science (AAA-DS) Program Award to the UvA and VU Universities. We would also like to thank the following organisations. The source data for the neighbourhoods task is published by Statistics Netherlands (CBS) and distributed by the Publieke Dienstverlening op de Kaart organization (PDOK) under a Creative Commons (CC) Attribution license. The data for the buildings task was published by the Dutch National Cadastre under a CC Zero license. The archaeological data in raw form is hosted by Data Archiving and Networked Services, and re-licensed by kind permission of copyright holder ADC ArcheoProjecten under CC-BY-4.0. We thank Henk Scholten, Frank van Harmelen, Xander Wilcke, Maurice de Kleijn, Jaap Boter, Chris Lucas, Eduardo Dias, Brian de Vogel and anonymous reviewers for their helpful comments.

\section{Introduction}
\label{sec:introduction}
For many tasks, it is useful to analyse the geometric shapes of geospatial objects, such as in quality assessment or enrichment of map data \citep{fan2014quality} or such as the classification of topographical objects \citep{keyes1999fourier}. Machine learning is increasingly used in geospatial analysis tasks. Machine learning can learn from data by extracting patterns \citep[2]{Goodfellow-et-al-2016}. For example, machine learning can be applied to classify building types \citep{xu2017quality}, analyse wildfires \citep{Araya2016}, traffic safety \citep{Effati2015}, cluster spatial objects \citep{Hagenauer2016}, detect aircraft shapes \citep{wu2016shape} or classify road sections \citep{Andrasik2016}: tasks that extend beyond standard GIS processing operations. The prediction of house prices \citep{Montero2018} and the estimation of pedestrian side walk widths \citep{Brezina2017} are tasks that could also benefit from the application of machine learning analysis on geometric shapes. 

Deep learning is a relatively new addition to the collection of machine learning methods. Deep learning allows stacking multiple learning layers to form a model that is able to train latent representations at varying levels of data abstraction \citep{lecun2015deep}. In this paper, we will use the term \emph{shallow} machine learning \citep[2-3]{ball2017comprehensive} to refer to methods that are \emph{not} based on deep learning methods. A distinguishing property of deep versus shallow learning methods is that shallow learning requires a preprocessing step known in machine learning as \emph{feature extraction} \citep[438]{lecun2015deep}, a lossy data transformation process. Shallow models require feature vectors as input data, so when using data of variable length such as geometries, shallow learning algorithms depend on feature extraction. One advantage of deep models over shallow models is that these feature extraction methods are not required for deep learning, which is why we want to explore the abilities of deep learning to operate on all available geometry data rather than on an extracted set of features. 

The purpose of this article is to assess the accuracy of working with vector geometries in deep neural nets, by comparing them with existing shallow machine learning methods in an experiment with three classification tasks on vector polygons. Our main objective is to train deep learning models on all available data. From this objective we do not require our deep learning models to exceed shallow model accuracy, but we do require the deep models to at least match shallow model accuracy. Thus, the main question we want to answer is: 
\begin{addmargin}[2em]{2em}
\indent \textbf{Can deep learning models achieve accuracies comparable with shallow learning models in analysing geospatial vector shapes?}
\end{addmargin}

These are the contributions made in this paper:
\begin{enumerate}
  \item We compare the performance of shallow and deep learning methods on geospatial vector data, as detailed in Section~\ref{sec:methods}. We show that the deep learning models introduced here match shallow models in accuracy at classification tasks on real-world geospatial polygon data. 
  \item We introduce three classification tasks restricted to geospatial vector polygons that serve as a novel and open access benchmark on geospatial vector shape recognition, detailed in Section~\ref{sec:tasks}. The benchmark data files are available as open data.\footnote{Data available at \url{http://hdl.handle.net/10411/GYPPBR}}
\end{enumerate}

Since the domains of geospatial information systems (GIS) and machine learning (ML) have partially overlapping vocabularies, we provide Table~\ref{tab:disambiguation} of homonyms and their use in the two fields of GIS and ML. Where used in this article, the terms are clarified by their field or, where possible, avoided.  

\begin{table}
\begin{tabular}{l p{4cm} p{4cm}} \toprule
 Term & GIS & ML \\ [0.5ex] \midrule
 Geometry & \raggedright A spatial representation of an object encoded as one or more points that may be interconnected & \\
 Vector & \raggedright A geometry defined by vertices and edges & A one-dimensional array \\ 
 Vectorization & \raggedright Conversion of raster or analog data 
 into geospatial vector geometries & Conversion of data into a tensor interpretable by a machine learning algorithm \\  
 Feature & A geospatial object & A data property \\
 Shape & \raggedright A geospatial object geometry & A tensor size along its dimensions \\
 K-nearest neighbours & \raggedright The $k$ spatially closest objects & A learning algorithm based on closest resemblance \\ \bottomrule
\end{tabular}
\caption{Terms in the fields of GIS and ML}
\label{tab:disambiguation}
\end{table}

\paragraph{Vector geometries as raster data}
Our research focuses on machine learning on geospatial vector data without rasterization of the geometry data. However, as any geometry can be expressed as raster data, the question must be addressed why one should not simply convert vector data to raster and use  machine learning algorithms that are commonly used on raster data. The answer to this is that geospatial vector data is often better suited for representation of discrete geospatial objects, because rasterization leads to loss of information almost everywhere: 

\begin{enumerate}
\item Geospatial vector data is highly versatile in representing geometries at varying spatial levels of detail, whereas raster data has a resolution of fixed and uniform size. Geospatial vector data can leverage these different levels to represent, for example, the rough shape of a country and the detailed shape of a microbe, at opposite sides of the globe within a single multi-part geometry.
\item Vector data is almost always more compact in comparison with raster data. Depending on the accuracy of the source data, materialisation of vector data into raster data often requires expansion to transform the vector data into a uniform rasterized sampling of a continuous field. 
\item Geospatial vector data can be reasoned over by any Geospatial Information System (GIS) in terms of topology: properties of geospatial objects with respect to other geospatial objects in the same set that are invariant under linear transformations, such as object intersection or spatial adjacency \citep[102]{huisman2009principles}. With rasterization, this information may be partially or completely lost: a small gap between two disjoint geometries for example may be lost if the pixel resolution is lower than the gap size. 
\end{enumerate}  

Thus, the rasterization process is trivial but lossy, where the inverse process of geospatial vectorization is non-trivial and requires human or algorithmic interpretation \citep[309]{huisman2009principles}. For these reasons, it is important to explore the capabilities of shape analysis by machine learning models without resorting to rasterization. 

The further article structure is as follows: we position our work within related research in Section~\ref{sec:related}, we explain the methods of our research in Section~\ref{sec:methods}, we discuss the classification tasks in Section~\ref{sec:tasks}, and the model performance results  on these tasks in Section~\ref{sec:results}.

\section{Related work}
\label{sec:related}
The vast majority of machine learning research in the geospatial domain is focused on analysis of remote sensing data, as shown by overview works from, for example, \citet{zhu2017} and \citet{ball2017comprehensive}; and by challenges such as the CrowdAI mapping challenge\footnote{\url{https://www.crowdai.org/challenges/mapping-challenge}} and the DeepGlobe Machine Vision Challenge\footnote{\url{http://www.grss-ieee.org/news/the-deepglobe-machine-vision-challenge/}} \citep{demir2018deepglobe}. 

Compared to remote sensing raster data, far fewer publications go into the matter of analysing geospatial \emph{vector} shape data through machine learning strategies. The most common method is to rasterize the vector shapes first. \citet{xu2017quality} have published a deep learning strategy for comparing building footprints. However, the approach by Xu et al. requires preprocessing that rasterizes aggregated data and does not classify individual geometries. Similarly, the shapes in the deep learning image retrieval task through sketches described by \citet{jiang2017} are raster-based rather than vector-based abstractions, as are the aircraft shapes extracted from remote sensing in the work by \citet{wu2016shape}. The work on 3D model retrieval by \citet{wang2017afeatureextraction} uses a different rasterization strategy: 2D-projected images are generated from 3D models to create an image search database. \citet[2-9]{cheng2016survey} survey a number of works involving geometric shape data, but aimed at classical (i.e. non-machine learning) remote sensing object detection strategies, rather than on machine learning analysis of the geometric shapes themselves. In contrast to the raster-based strategies from these works, we aim to research the possibility of avoiding the rasterization process and operate on geometries directly, as will be explained in Section~\ref{sec:preprocessing}.

Research on machine learning analysis of non-rasterized vector shapes is scarce. The algorithms used by \citet{Andrasik2016} are trained directly on geometry properties based on angles and radii of vertices in road sections, extracted from simplified road geometries. Their method, however, is optimized to the specific task of classifying short road sections from short polylines. \citet[120-121]{Effati2015} adopt a similar strategy for the road properties for the purposes of traffic safety analysis. We aim to explore more generic shape analysis methods through machine learning, rather than task-specific ones. A deep learning model operating on vector geometries was developed by \citet{ha2018a}, using a model they named \emph{sketch-rnn}. Sketch-rnn shows how a deep learning architecture can be used work with vector geometries directly. The data collected for sketch-rnn used a web-based crowd-sourcing tool, inviting users to draw simple vector drawings of cats, t-shirts and a host of other object categories. Given an object category, the generative sketch-rnn model is able to analyse partial shapes drawn by the user and extrapolate these to complete sketches.\footnote{\url{https://magenta.tensorflow.org/assets/sketch_rnn_demo/index.html}} 
\section{Methods}
\label{sec:methods}
The classification tasks in this paper operate on real-world polygon data. To be precise, we use the term \emph{polygon} to mean a single connected sequence (i.e. without polygon holes) of three or more coplanar lines. Every line in a polygon is defined by two points in $\mathbb{R}^2$, where each point is shared by exactly two lines to form a closed loop. We impose no validity constraint on polygons, i.e. polygons may be self-intersecting. 

\subsection{Shallow models}
\subsubsection{Preprocessing}
Shallow machine learning methods operate on feature vectors of fixed length, so more complex input data such as geometries need to be transformed. This transformation step is known in machine learning as \emph{feature extraction} \citep[438]{lecun2015deep} or \emph{feature engineering} \citep[84]{domingos2012few}. So, contrary to deep learning models discussed in Section~\ref{sec:deep_models}, shallow methods do not operate on geometry coordinates directly, since vector geometries are sequences of vertices that are both of higher rank and of variable length, as we will explain in Section~\ref{sec:preprocessing}. Applied to geospatial vector data, we rely on extracting information from a geometry that characterizes its shape in a lower-dimensional, fixed-length representation. 

\begin{figure}
  \includegraphics[width=1\textwidth]{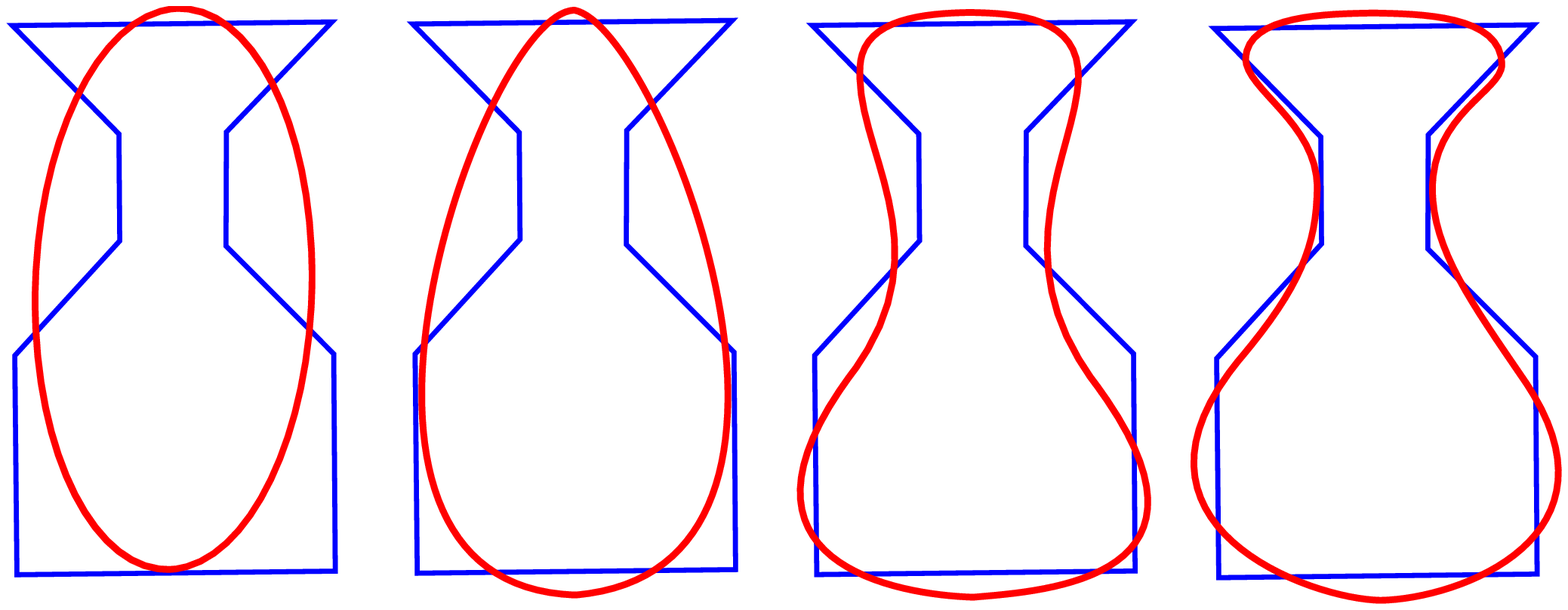}
  \caption{Order 1, 2, 3 and 4 elliptic Fourier reconstruction approximations (red) of a polygon (blue). Each order level adds to the approximation. Adapted from \citet[237]{kuhl1982elliptic}}
  \label{fig:efd_orders}
\end{figure}

Fourier descriptors are a common choice as a feature engineering method for extracting properties from geometries \citep{zhang2002comparative,keyes1999fourier,zahn1972fourier,LONCARIC1998983}. For our preprocessing, we used the Elliptic Fourier Descriptor (EFD) method by \citet{kuhl1982elliptic}. Elliptic Fourier descriptors are created by iterating over the coordinates of the vertices in a geometry, transforming any number of coordinates of the geometry into a vector representing the geometry in an elliptic approximation. This transformation can be reversed, producing an approximation of the original geometry, its reconstructive accuracy depending on the \emph{order} or the number of \emph{harmonics} \cite[239]{kuhl1982elliptic}, the order or number being a positive integer. The higher the order, the better the approximation gets, as shown in Figure~\ref{fig:efd_orders}. 

The Fourier descriptors were constructed using the \emph{pyefd} package,\footnote{https://pypi.python.org/pypi/pyefd} which implements the algorithms by \citet{kuhl1982elliptic} made specifically for creating descriptors for vector geometries. The pyefd package produces normalized and non-normalized descriptors; the normalized descriptors are start position, scale, rotation and translation invariant \cite[236]{kuhl1982elliptic}. For the data used in training the shallow models, both normalized and non-normalized Fourier descriptors were included. 
Added to the descriptors are three other easily obtained geometry properties: the polygon surface area, number of vertices and geometry boundary length.  

\subsubsection{Shallow model selection}
As explained in Section~\ref{sec:introduction} we distinguish between two families of machine learning methods. From the shallow model family, we selected four standard algorithms: 
\begin{itemize}
\item K-nearest neighbour classifier;
\item Logistic regression;
\item Support Vector Machine (SVM) with Radial Basis Function (RBF) kernel. Other kernels (linear, polynomial) were tested but RBF always produced better results;
\item Decision tree classifier.
\end{itemize}

These model types are so well-established that we will not elaborate upon these here. Logistic regression goes back to \citet{cox1958}, decision trees to \citet{breiman1984}, k-nearest neighbour to at least \citet{cover1967nearest}. With over twenty-five years of history, the SVM by \citet{boser1992training} is the relatively new algorithm in the shallow model family.

\subsection{Deep models}
\subsubsection{Preprocessing}
\label{sec:preprocessing}
The problem with shallow model feature extraction of geospatial vector data is that it results in information loss: the original shapes can only be approximated, but not fully reconstructed by reversing the feature extraction process. Ideally, it would not be necessary to extract an information subset of geospatial data in advance to obtain good predictions. Deep learning allows us to train models on geospatial vector data by directly feeding the geometry coordinates to a deep learning model, without extracting intermediate features. To explain how deep learning is able to learn from polygon data directly, we need to discuss our machine learning vectorization method. 

\subsubsection{Geometries as machine learning vector sequences}
\begin{figure} \centering
	\begin{subfigure}[b]{\textwidth}
		\includegraphics{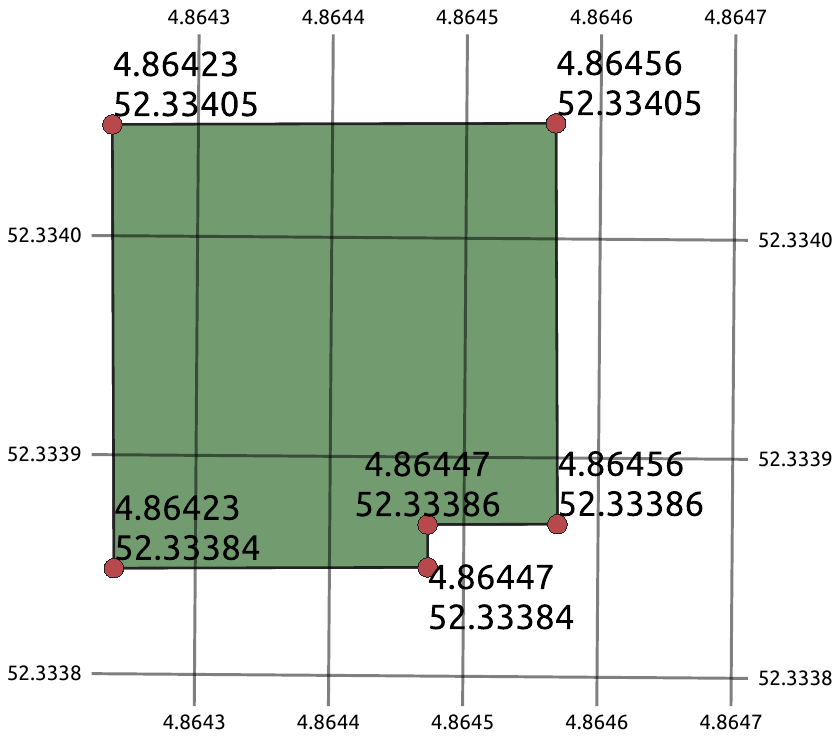}
        \label{fig:building_map_sample}
        \caption{}
	\end{subfigure}
    
	\begin{subfigure}[b]{\textwidth}
    \begin{tabular}{p{.3\textwidth} p{.3\textwidth} p{.3\textwidth}}
      Polygon coordinates & Center: remove mean of [4.8644271, 52.3339057] & Scale: divide by  scale factor of 2.64501e-4 \\ \toprule 
4.86447, 52.33384 & 4.2857e-5, -6.5714e-5 & 0.16198, -0.24845 \\ 
4.86447, 52.33386 & 4.2857e-5, -4.5714e-5 & 0.16198, -0.17283\\
4.86456, 52.33386 & 1.32857e-4, -4.5714e-5 & 0.50229, -0.17283\\
4.86456, 52.33386 & 1.32857e-4, 1.44286e-4 & 0.50229, 0.54550\\
4.86423, 52.33405 & -1.97143e-4, 1.44286e-4 & -0.74534, 0.54550\\
4.86423, 52.33405 & -1.97143e-4, -6.5714e-5 & -0.74534, -0.24845\\
4.86447, 52.33384 & 4.2857e-5, -6.5714e-5 & 0.16959, -0.24845\\
	  \end{tabular}
      \caption{}
	\end{subfigure}
    
	\begin{subfigure}[b]{\textwidth}
      \begin{tabular}{l}
        Tensor representation\\ \toprule
        {[}0.16198, -0.24845,  1, 0, 0],\\
        {[}0.16198, -0.17283,  1, 0, 0], \\
        {[}0.50229, -0.17283,  1, 0, 0], \\
        {[}0.50229, 0.54550,   1, 0, 0], \\
        {[}-0.74534, 0.54550,  1, 0, 0], \\
        {[}-0.74534, -0.24845, 1, 0, 0], \\
        {[}0.16959, -0.24845,  0, 0, 1]  \\
      \end{tabular}
      \caption{}
	\end{subfigure}
    
  \caption{A building polygon (a) with its coordinate normalization by local mean subtraction and global scaling (b) and the vector representation (c). Coordinates (in CRS84 projection) and standard deviation have been truncated to five digit precision for the sake of brevity. In the final tensor representation (c) the render type is added.}
  \label{fig:geom2vector}
\end{figure}

Our geometry tensor encoding was derived from the work by \citet{ha2018a}, where each geometry sample $G_i$ in a data set of size $n$ is encoded as a sequence of geometry vertex vectors:  
$\langle
	\mathbf{g_1^i}, \mathbf{g_2^i}, \mathbf{g_3^i}, \ldots, \mathbf{g_m^i}
\rangle$
, and where $m$ is the number of vertices in the geometry. 
Each vector $\mathbf{g_j^i}$ is a concatenation of: 
\begin{itemize}
\item a coordinate point vector $\mathbf{p_j^i}$ in $\mathbb{R}^2$. In fact any coordinate system in $\mathbb{R}^2$ is supported in this vector representation.
\item A one-hot vector $\mathbf{r_j^i}$ in $\mathbb{R}^3$ to mark the end of either the point, sub-geometry or a final stop for the vertices in $G^i$. For each $\mathbf{g_j^i}$ in a polygon geometry, 
$\mathbf{r_j^i} = \begin{bmatrix} 1 & 0 & 0 \end{bmatrix}$ except for the last vertex, where $\mathbf{r_m^i}$ marks the end of the polygon as the final stop $\begin{bmatrix} 0 & 0 & 1 \end{bmatrix}$. In case of a multipolygon, each sub-polygon is terminated by a sub-geometry stop $\begin{bmatrix} 0 & 1 & 0 \end{bmatrix}$ except for the last, which is marked as a final stop.
\end{itemize}
Combined, $\mathbf{g_j^i}$ is a vector of length 5, as shown in  Figure~\ref{fig:geom2vector}. 

As we explained in Section~\ref{sec:introduction}, the rasterization process is lossy. As such, it can be considered a feature extraction method by which we lose information that may be of use to the machine learning model. So, for our deep neural net architectures, geometries are expressed as normalized vector sequences of geometry vertices. This process is fully reversible in reconstructing the geometry shape and orientation, but it does require centering and scaling the data. For deep learning models to perform, the data needs to be normalized to a mean of zero and scaled to a variance of about one.  

Geospatial coordinates are often expressed in degrees of longitude and latitude, where one degree of latitude equals roughly 111 kilometres. However, vector geometries usually operate on the level of meters or even centimetres. To counter this normalization imbalance, every point vector $\mathbf{p_j^i}$ in every geometry $G^i$, $\mathbf{p_j^i}$ is normalized to
\begin{equation}
  \mathbf{p{_j^i}'} = 
  \frac{\mathbf{p_j^i - \mathbf{\overline{p^i}}}}
  {s},
\end{equation}
where $\mathbf{\overline{p^i}}$ is the geometry centroid of geometry $G^i$, computed as the mean average of all $\mathbf{p^i}$ in a \emph{single} geometry $G^i$. Scale factor $s$ is the standard deviation over the bounding values $b^i_{min}$ and $b^i_{max}$ of \emph{all} geometries. $b^i_{min}$ and $b^i_{max}$ for a geometry $G^i$ are defined as
\begin{equation}
	b^i_{min} = \min\left(\mathbf{p^i} - \mathbf{\overline{p^i}}\right), 
\end{equation} and 
\begin{equation}
	b^i_{max} = \max\left(\mathbf{p^i} - \mathbf{\overline{p^i}}\right).
\end{equation}
This is a simpler two-value version of the standard bounding box that would normally list the minimum and maximum values for a geometry in two dimensions. Scale factor $s$ is then computed as the scalar standard deviation over all bounding values $B$:
\begin{equation}
 B = \langle b^1_{min}, b^1_{max}, b^2_{min},b^2_{max}, \ldots, b^n_{min},b^n_{max}\rangle
\end{equation}

Geometries often vary in the number of vertices required to approximate the shape of a real-world object. As a consequence, the geometry vector sequences vary in length. Deep learning models have the benefit of being able to train and predict on variable length sequences \citep{DBLP:journals/corr/BahdanauCB14}. However, within one batch the sequences need to be of the same length in order to uniformly apply the model weights and biases on the entire batch as a single tensor. To achieve this fixed sequence size within a batch, the geometry vectors are first sorted in reverse order, with the largest geometry first and the smallest last. This sorted set of geometries is subdivided into bins of size $n_{bin}$, where $n_{bin}$ is at least the training batch size. This is to increase computational efficiency and reduce training time on what otherwise would be a large array of very small batches. If there are insufficient geometries of sequence length $m_{bin}$ to create a set of samples of batch size, smaller geometries are added and padded to sequence length $m_{bin}$. Thus, a geometry with a sequence length $m$ of 144 points is zero-padded to a size $m_{bin}$ of 148 if the largest sequence length in the batch is 148. This preprocessing of binning and limited padding reduced the training time to one quarter of the time needed for training on fixed size sequences.

Although there is no theoretical upper bound to the sequence length, there is a practical one for the amount of memory on commodity hardware. The data sets contain a small amount of very large geometries. To improve computational efficiency and prevent memory errors, these rare cases are simplified using the Douglas-Peucker algorithm \citep{douglas1973algorithms}. In this way, only 0.17 percent  of the geometries in our experiments needed to be simplified.  
% simplified archaeo: 190 out of 56592
% simplified buildings: 16 out of 160451
% simplified neighbourhoods: 195 out of 13208
% total 401 out of 230251 = 0.001741578

\subsection{Deep model selection}
\label{sec:deep_models}
The motivation for using deep learning on geospatial vector data goes beyond matching or improving existing methods. Deep learning allows us to explore new methods for working with geospatial data, in complex pipelines involving combinations of raster, numerical and textual data \citep{ngiam2011multimodal}, including geospatial vector data. Deep learning can be used for classification or regression tasks, but also for training generative models, producing new text \citep{sutskever2014sequence}, image \citep{NIPS2014_5423} and even vector shape \citep{ha2018a} outputs. Knowing how well deep learning models can learn directly from geometries is a first step in building more complex generative pipelines with confidence that the model is able to correctly interpret the data. 

We introduce two deep learning, end-to-end trained models where vector-serialized geometries are given as input data. The deep learning model figures out the relevant data properties for itself without the need for the feature extraction required for the shallow models. In the next subsections we describe two relatively simple deep learning architectures that we evaluate on our tasks: a convolutional model and a recurrent model. We explain these deep learning models in some detail. 

\begin{figure}
  \begin{minipage}[t]{.5\textwidth}
	\vspace{0pt}
    \centering
    \includegraphics[width=1\textwidth]{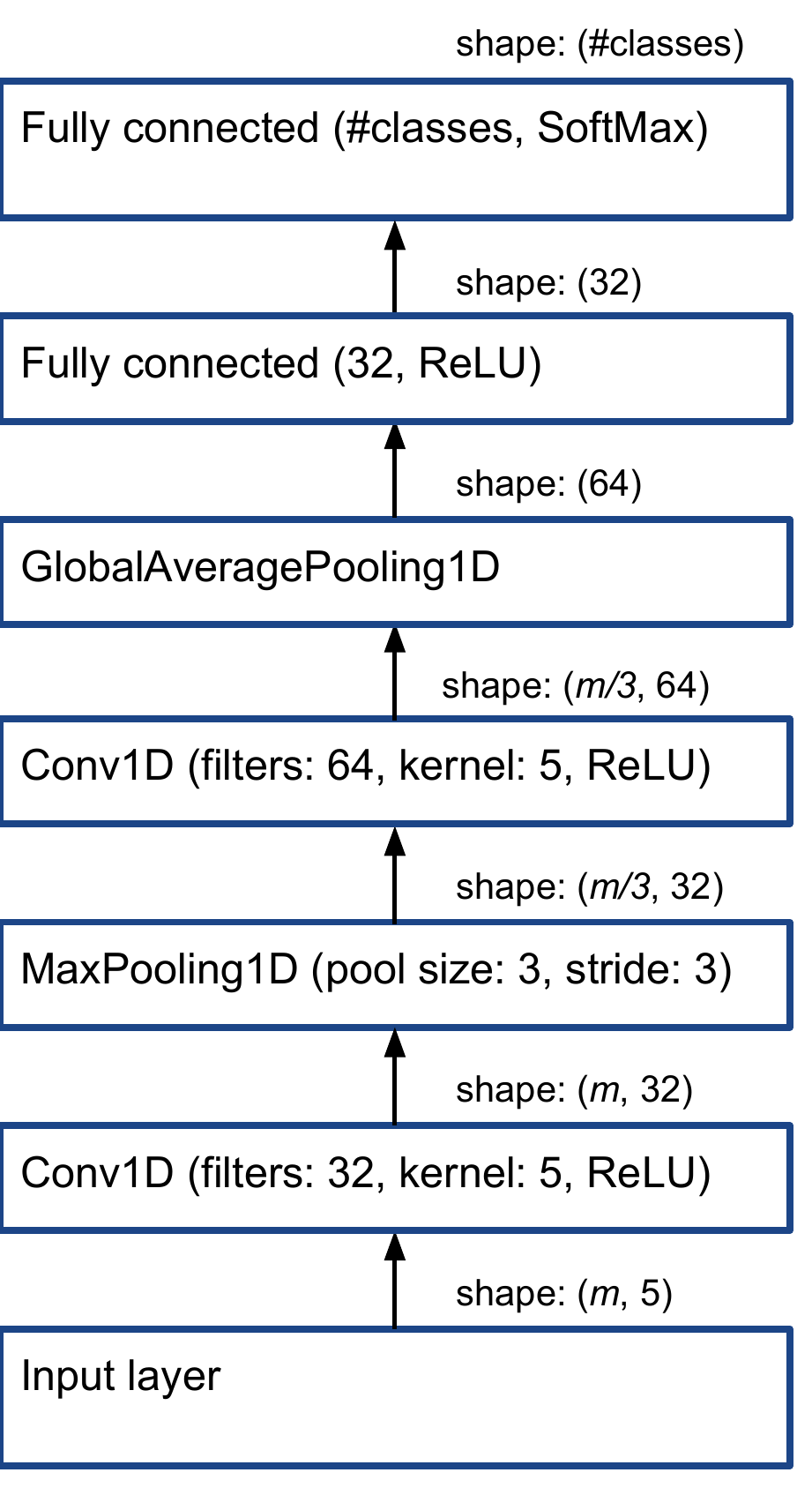}
  \end{minipage}
  \begin{minipage}[t]{.5\textwidth}
	\vspace{0pt}
    \centering
    \includegraphics[width=1\textwidth]{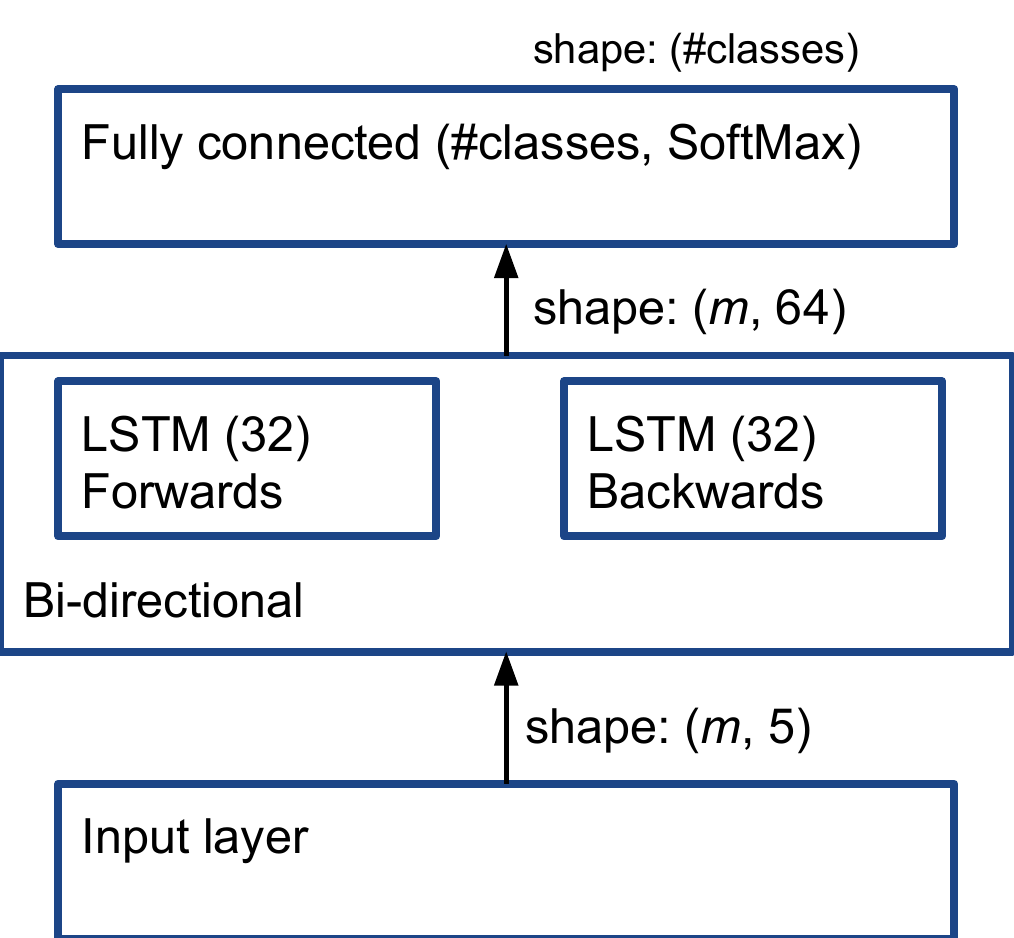}
  \end{minipage}
  \caption{Convolutional (left) and recurrent (right) model layouts.}
  \label{fig:deep_models}
\end{figure}

\subsubsection{Convolutional neural net}
\begin{figure}
    \includegraphics[width=1\textwidth]{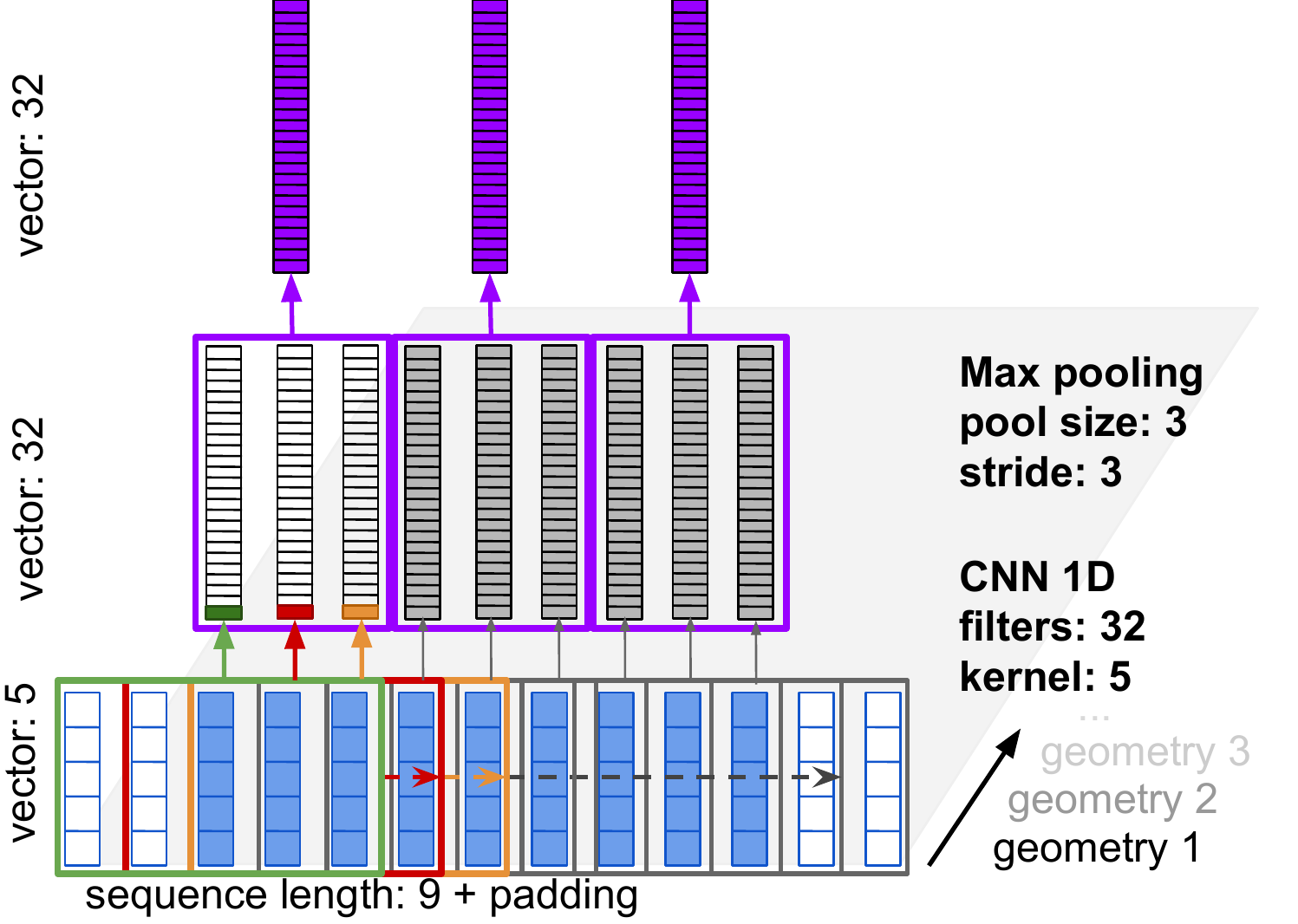}
    \caption{The first two layers of the CNN model. With a kernel size of five, the CNN inspects a sliding window over the first five geometry vectors in geometry $G^1$, producing the green element in the CNN output vector. The CNN then moves to the five elements  to the right, and produces the red vector element, next the orange vector, repeated until the end of the geometry (the next three windows in grey). This process is repeated for each filter and then moves to the next geometry, in the direction of the black arrow. The max pooling operation combines the maximum output element values of the CNN, shown in purple for geometry 1.}
    \label{fig:convnet_maxpool_architecture}
\end{figure}

The first introduced deep learning model uses a 1D convolutional neural net (CNN) layout, shown in Figure~\ref{fig:deep_models}. For an introduction to the workings of the CNN, we refer the reader to \citet{olahconvnets}. As a first layer, our model uses a ReLU-activated convolution layer with a filter size of 32, a kernel size of five and a stride of one. With this configuration, the CNN starts a sliding window across the first five geometry vectors, i.e. $g_1^i$ through $g_5^i$, producing a vector of size 32 as specified by the filter hyperparameter\footnote{Hyperparameters are the configuration settings of the machine learning model that are not optimized by the algorithm itself, such as the batch size.}. This window of size five is slid along the vectors of the geometry, until the end of the geometry including padding. Padding ensures outputs by the CNN of the same sequence length as the input, to prevent size errors on small geometries where the tensor size becomes too small to pass through the specified network layers. After $g_1^i$ through $g_5^i$ the CNN continues at the second set of geometry entries $g_2^i$ through $g_6^i$. After inspecting all values of all the vectors in the first geometry, the CNN continues at the next geometry (see Figure~\ref{fig:convnet_maxpool_architecture}).

\begin{figure}
    \includegraphics[width=1\textwidth]{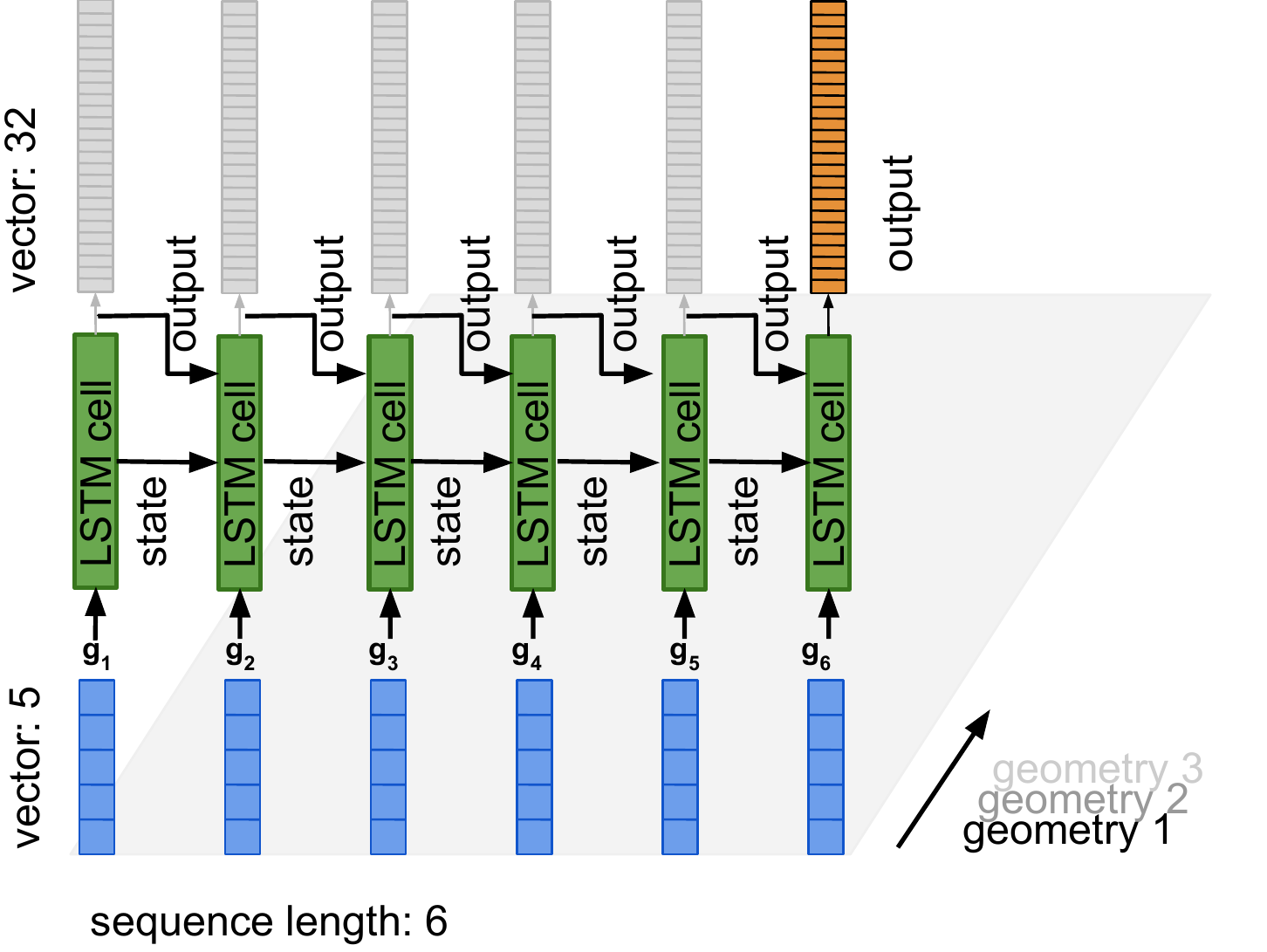}
    \caption{Forward-facing LSTM, as part of the first layer of the LSTM model. Unlike the CNN architecture, complete vectors are fed one by one to the same LSTM cell. The green boxes therefore represent the same cell, with only its state updated: along with each next geometry vector, the output and previous state of the LSTM cell are passed along from one vector to the next. For the purposes of classification as in this article, only the last LSTM output is returned (in orange), the intermediate outputs (in grey) are discarded.}
    \label{fig:lstm_architecture}
\end{figure}
The first CNN layer is followed by a max pooling layer with a pooling size of three and a stride of three. The max pooling operation with a pool size of three combines the maximum values of three CNN output vectors into a single sequence vector of the same length. The reduction of the CNN output to one-third is specified by the max pooling \emph{stride} hyperparameter: after combining CNN output vectors $\mathbf{c_1^i}$, $\mathbf{c_2^i}$ and $\mathbf{c_3^i}$, the max pooling operation skips forward to combine outputs $\mathbf{c_4^i}$, $\mathbf{c_5^i}$ and $\mathbf{c_6^i}$, and so on. After the max pooling layer, a second convolution layer (not shown in Figure~\ref{fig:convnet_maxpool_architecture}) interprets the output of the max pooling layer, with hyperparameters identical to the first but with 64 filters instead of 32. This CNN layer is followed by a global average pooling layer that reduces the tensor rank to two by computing the average over the third tensor axis. The output is subsequently fed to a ReLU activated fully connected layer. The last layer is a softmax-activated fully connected layer to produce probability outputs that sum to one.

\subsubsection{Recurrent neural net}
\label{sec:rnn}
The second deep learning model uses a recurrent neural net (RNN) layout as shown in Figure~\ref{fig:deep_models}. An RNN processes input data sequentially, re-using the outputs of the network in each slice of the input sequence as additional input for the next slice \citep[364]{Goodfellow-et-al-2016} by which the network can carry over information between slices and capture long-term dependencies in the input sequence. While there are several RNN architectures, we opted to evaluate an Long-Short Term Memory (LSTM) \citep{hochreiter1997long} architecture, as this is one popular architecture, shown by \citet{ha2018a} to be effective for vector geometry analysis. 

The core of our model is a single bi-directional LSTM layer. The LSTM architecture is a particular type of RNN, designed to process sequences of data with a trainable \emph{forget gate}. This forget gate regulates the information retained from one geometry vertex to the next. During training, the LSTM learns which information in the sequence is of interest to retain, by passing both the vector in the sequence, the output and the cell state from one input vector to the next (see Figure~\ref{fig:lstm_architecture}). For a detailed discussion of recurrent neural nets and LSTMs in the geospatial domain, we refer the reader to \citet{7914752}. An introduction to LSTMs is given by \citet{olahlstms}. LSTMs have been shown to be effective on sequences such as words in a sentence \citep{sutskever2014sequence}, but also sequential geometry-like data such as handwriting recognition and synthesis \citep{DBLP:journals/corr/Graves13}. The ability of the LSTM to learn long-term dependencies \citep[400]{Goodfellow-et-al-2016} in sequences of input data renders it a suitable architecture to test its abilities on geospatial vector geometries. 

The bi-directional architecture feeds the sequence of geometry vertices forwards as well as backwards through LSTM cells \citep{650093}, where the resulting output from these cells is concatenated. This allows the network to learn from the preceding vertices in the geometry as well as the ones that are ahead. In our model, we configured both the forwards and backwards LSTM to produce an output of size 32, combined to 64. As with the CNN model, the last layer is a softmax-activated fully connected layer to produce probability outputs that sum to one.

\subsection{Evaluation methodology}
\label{sec:evaluation_methodology}
To compare our selected shallow and deep learning models, we designed a set of experiments applicable to both shallow and deep models, informative to our research question and straightforward to interpret. We focus on a set of classification tasks as explained in Section~\ref{sec:tasks}, for which our models are required to correctly assign individual polygons to a certain type, based on only the polygon shape. 

For the classification tasks, we select data sets on the following requirements: 
\begin{itemize}
\item Each task contributes to evaluate the accuracy performance of deep learning models versus shallow models. The data sets for the tasks contain real-world polygon data from different domains, with different use cases and on different spatial scales;
\item Each data set contains enough data to draw conclusions on model generalization; we set a requirement for data sets of at least 12,000 geometries in order to provide a training set of at least 10,000 geometries, a validation set of at least 1,000 geometries and a test set of at least 1,000 geometries; 
\item Each task requires the models to infer information from the polygon shape, and from the polygon shape alone. To this end, data is selected to be likely to contain shape information relevant to the classification task but not a trivial solution;
\item We require our data to be available under an open license, to be accessible for future research.
\end{itemize}
Through the use of classification tasks, both the shallow and deep model performance can be directly compared. Classification tasks can be expressed through the simple metric of accuracy score: the ratio of correctly assigned test samples over the total set of test samples. We also add a baseline majority class accuracy score: the fraction of the prevalent class, included as a baseline of the most simple method to exceed. The models are trained on a training data set, the model performance was iteratively tuned to best perform on an evaluation set and finally tested once on a test set that was unseen during any of the tuning runs. The resulting test set accuracy scores are listed in Table~\ref{tab:evaluation}. We include a discussion of the misclassification behaviour of the models by analysing the confusion matrices \citep{stehman1997selecting}.

For optimal reproducibility, we use only open data\footnote{Data available at \url{http://hdl.handle.net/10411/GYPPBR}} and open source methods to answer our research question. We release all preprocessing code, deep learning models and shallow models as open source software.\footnote{Code available at \url{https://github.com/SPINlab/geometry-learning}} Scikit-learn \citep{scikit-learn} provides the shallow learning algorithms. To evaluate shallow model accuracy on each task, we use a brute force grid search with 5-fold cross validation to find the best applicable hyperparameters for k (k-nearest neighbours), degree (decision tree), C (SVM, logistic regression) and gamma (SVM). SVM grid searches are restricted to a maximum number of 10M iterations to allow the grid search operation to complete within a day. Grid searches on SVM models and k-nearest neighbours are restricted to a subset of the training data to allow the grid search to finish within a day on commodity hardware. All shallow models are trained, however, using the full training set on the the best hyperparameters obtained from the grid search. The deep learning models are implemented using Keras \citep{chollet2015keras} version 2 with a TensorFlow \citep{DBLP:journals/corr/AbadiABBCCCDDDG16} version 1.7 backend. All deep model hyperparameters are tuned on the full training and validation set, using validation data split randomly from the training data. 

For the grid searches, we do not assume that including an arbitrarily high number of descriptors produces the best accuracy score. Instead, the number of extracted Fourier descriptors used during training is included as a hyperparameter in the grid search for each shallow model, to produce the descriptor order at which the grid search obtains the best results. The best parameters found in the grid searches are listed in Table~\ref{tab:baseline_hyperparameters} of Appendix~\ref{sec:baseline_hyperparams}. 

\section{Tasks}
\label{sec:tasks}
We created a set of three classification tasks to evaluate the performance of several machine learning algorithms in shape recognition on real-world polygon data. From the requirements listed in Section~\ref{sec:evaluation_methodology} the following tasks and accompanying data sets were selected:
\begin{enumerate}
\item Predicting whether the number of inhabitants in a neighbourhood is above or below the national median, based on the neighbourhood geometry;
\item Predicting a building class from the building contour polygon. The available classes are buildings for the purpose of gathering; industrial activity; lodging; habitation; shopping; office buildings; health care; education; and sports.
\item Predicting an archaeological feature type from its geometry. Features are available as an instance of either a layer; wall; ditch; pit; natural phenomenon; post hole; well; post hole with visible post; wooden object; or recent disturbance.
\end{enumerate}
The classes and their frequencies are displayed in Table~\ref{tab:tasks}.  

\begin{figure}
%   \begin{minipage}[t]{1\textwidth}
% 	\vspace{0pt}
%     \centering
	\includegraphics{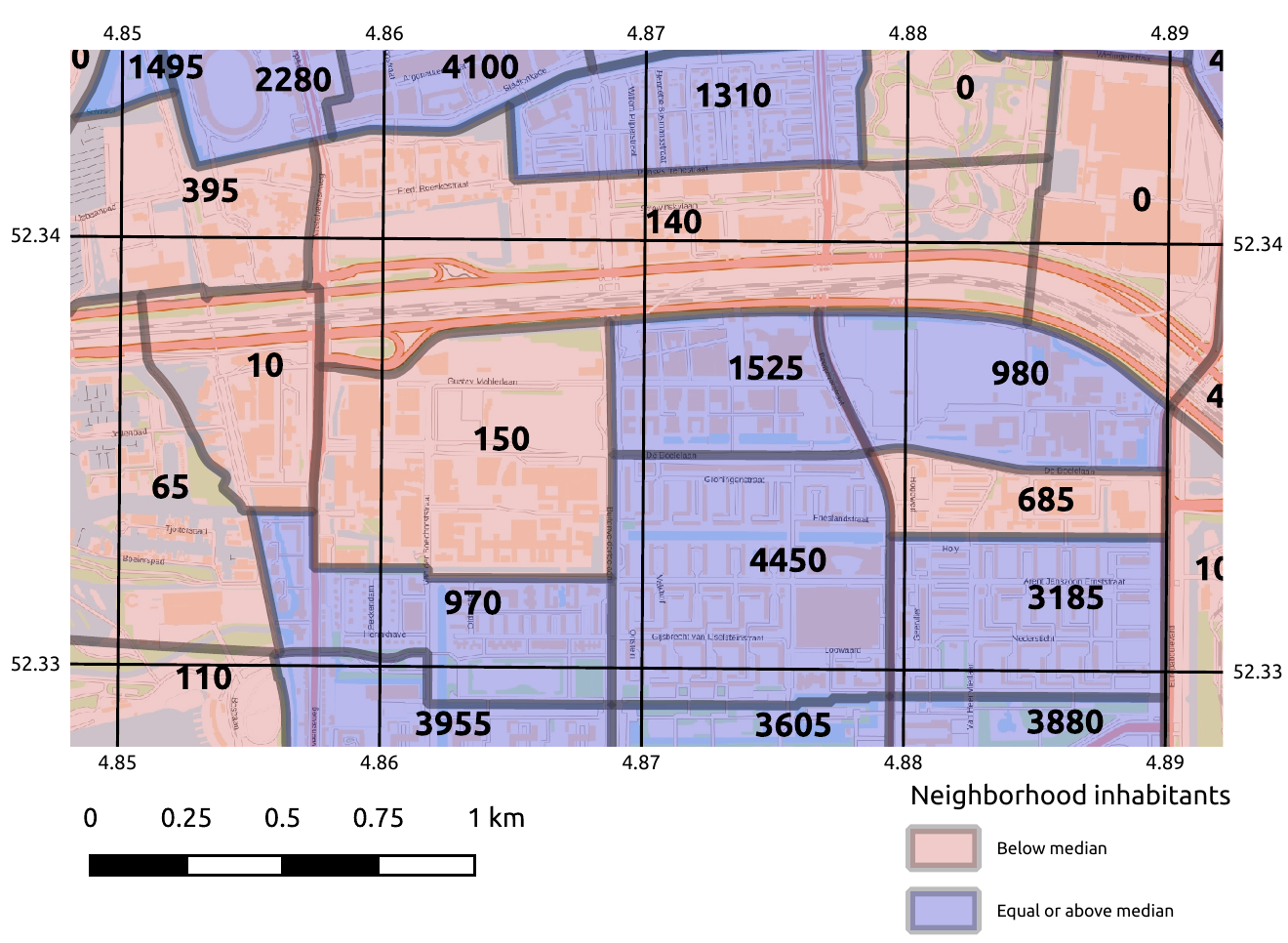}
	\label{fig:neighbourhoods}
	\caption{Map visualisation of neighbourhoods at scale 1:25,000 in Amsterdam near the VU University complex. In bold the absolute number of inhabitants per neighbourhood. Zero-count neighbourhoods are areas such as parks or office zones. Map coordinates in WGS84 degrees.}
%   \end{minipage}
\end{figure}
\begin{figure}
%   \begin{minipage}[t]{1\textwidth}
% 	\vspace{0pt}
%     \centering
    \includegraphics{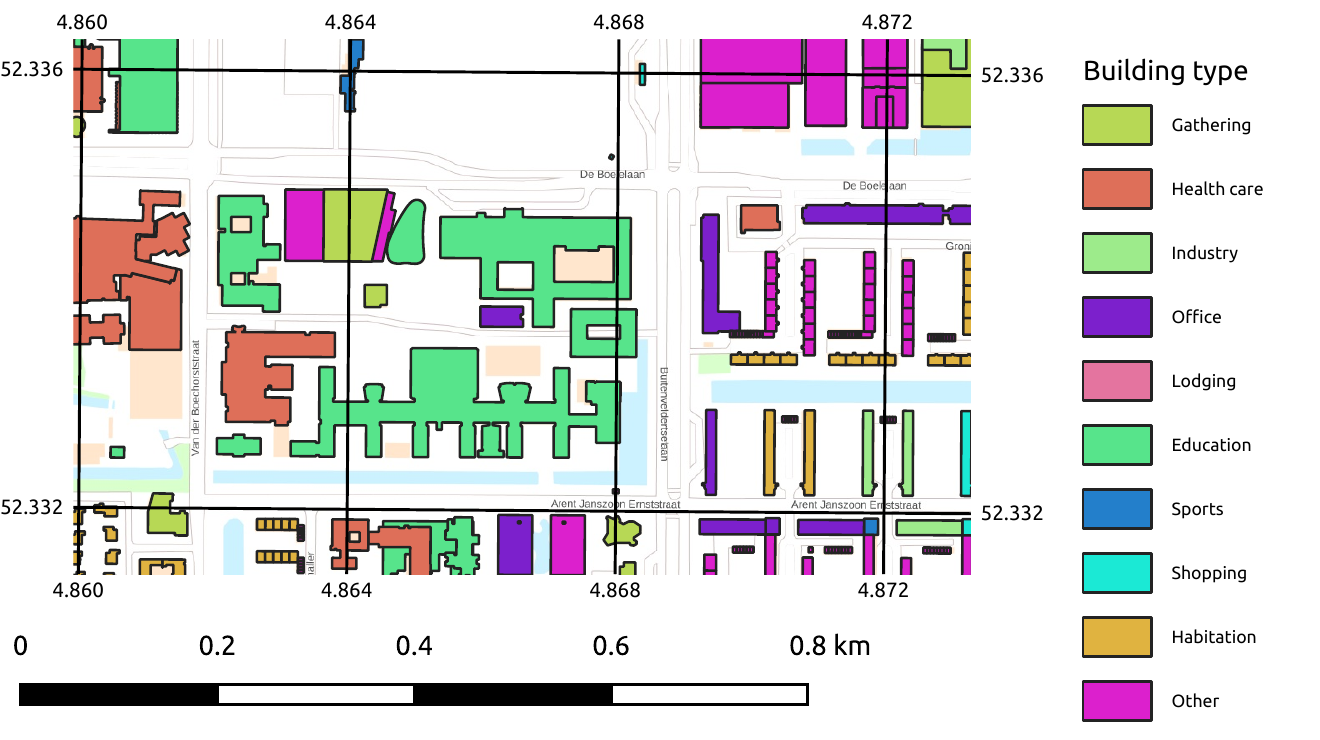}
	\label{fig:buildings}
	\caption{Map visualisation of buildings at scale 1:10,000 near the VU University complex. Map coordinates in WGS84 degrees.}
%   \end{minipage}
\end{figure}
\begin{figure}
%   \begin{minipage}[t]{1\textwidth}
% 	\vspace{0pt}
%     \centering
    \includegraphics{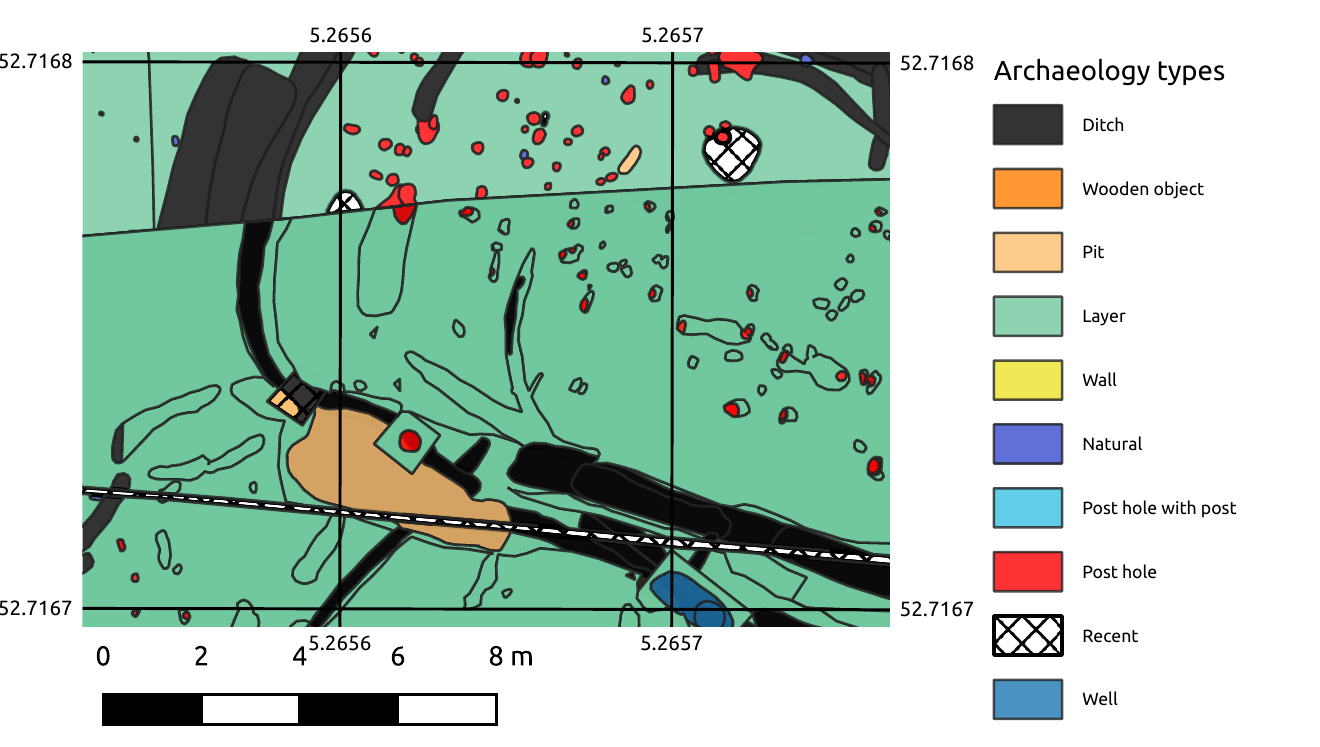}
%   \end{minipage}
  \caption{Map visualisation of archaeological ground features at scale 1:500, from \citep{ENKN-09}. Map coordinates in WGS84 degrees.}
  \label{fig:archaeology}
\end{figure}

\begin{table}
  \begin{minipage}[t]{.3\textwidth}
	\vspace{0pt}
    \begin{tabular}{ l r }
      \multicolumn{2}{p{.9\textwidth}}{Neighbourhood \linebreak inhabitants} \\ \toprule
      Class 		& frequency \\ \midrule
      $\geq$ median	& 6,610 \\
      $<$ median 	& 6,598 \\ \bottomrule
      Total 		& 13,208
    \end{tabular}
  \end{minipage}
  \begin{minipage}[t]{.05\textwidth}
  \end{minipage}
  \begin{minipage}[t]{.3\textwidth}
    \vspace{0pt}
    \begin{tabular}{ l r }
    \multicolumn{2}{p{.9\textwidth}}{Buildings\linebreak} \\ \toprule
    Function & frequency \\ \midrule
    Habitation 		& 23,000 \\ 
    Industrial 		& 23,000 \\
    Lodging 		& 23,000 \\
    Shopping 		& 23,000 \\
    Gatherings 		& 22,007 \\
    Office 			& 21,014 \\
    Education 		& 10,717 \\
    Healthcare 		& 7,832 \\
    Sports 		 	& 6,916 \\
    \bottomrule
    Total & 160,486
    \end{tabular}
  \end{minipage}
  \begin{minipage}[t]{.05\textwidth}
  \end{minipage}
  \begin{minipage}[t]{.3\textwidth}
    \vspace{0pt}
    \begin{tabular}{ p{.6\textwidth} r }
      \multicolumn{2}{p{.9\textwidth}}{Archaeological \linebreak features} \\ \toprule
      Class	& frequency \\ \midrule
      Post hole 							& 24,991 \\
      Pit 									& 7,713  \\
      Natural \linebreak phenomenon			& 6,136  \\
      Recent \linebreak disturbance			& 5,625  \\
      Ditch 								& 4,926  \\
      Wooden object 						& 2,499  \\
      Layer 								& 1,321  \\
      Wall 									& 1,387  \\
      Post hole with \linebreak visible post & 1,005 \\
      Water well 							& 980 \\ \bottomrule
      Total & 56,583
    \end{tabular}
  \end{minipage}
  \caption{Class frequency for the three tasks of neighbourhood inhabitants (left), building types (middle) and archaeological feature types (right).}
  \label{tab:tasks}
\end{table}

\subsection{Neighbourhood inhabitants}
The first task is to predict the number of inhabitants for a certain neighbourhood to be above or below the Dutch national median of 735 inhabitants. A neighbourhood is a geographical region of varying size and shape, as defined by Statistics Netherlands.\footnote{\url{https://www.cbs.nl/-/media/\_pdf/2017/36/2017ep37\%20toelichting\%20wijk\%20en\%20buurtkaart\%202017.pdf} (only available in Dutch).} The data was harvested from the 2017 districts and neighbourhoods Web Feature Service.\footnote{\url{https://geodata.nationaalgeoregister.nl/wijkenbuurten2017/wfs}} For the sake of evaluation simplicity, the task has been shaped into a binary class prediction: to predict a neighbourhood for having equal or more (6,610 neighbourhoods) or less (6,598 neighbourhoods) than the median of the absolute number of inhabitants of the entire set of neighbourhoods for the Netherlands for the year 2017. The median was chosen to create an even split\footnote{The slight difference between the class frequencies is explained by a higher occurrence of neighbourhoods with number of inhabitants equal to the median number.} of the two classes. For purposes of evaluation simplicity it is appropriate to estimate the median on the absolute number of inhabitants rather than population density. Calculating population density would add complexity to the task by requiring the models to divide the estimated number of inhabitants over an estimate of the neighbourhood surface area. Designed as a binary classification task, its purpose is primarily to evaluate model performance. However, the task could be re-designed to solve real-world problems: for example in estimating population sizes or densities for disaster areas based on topographical region shapes \citep{brown2001rapid}.  

\subsection{Building types}
The second task is to classify a building from the building footprint geometry, harvested from the buildings and addresses base registration (BAG) Web Feature Service.\footnote{\url{https://geodata.nationaalgeoregister.nl/bag/wfs}} The data set consists of buildings in nine functional classes, for a total of 160,486 buildings in the combined training and test dataset. Since the complete source data set comprises over five million buildings, each class was trimmed to a maximum of the first 23,000 instances per class to prevent creating a data set too large to experiment on. With this selection, the buildings set still has the most data of the three tasks. The task of classifying a building based on its shape can have clear quality control benefits: one could assess the likelihood of a specific building to belong to a specific type based on shape characteristics. Subsequently, a system could flag buildings as suspect that fall below a likelihood threshold for their current building class and suggest a replacement building class based on a higher likelihood.

\subsection{Archaeological features}
The third task is to classify an archaeological feature from its observed geometry. Archaeological features are field observations of disturbances in the subsoil as a result of human activities in the past. Archaeological institutions in the Netherlands store the results of archaeological field research in a digital repository \citep{gilissen2017archiving,hollander2014depot}. From the DANS EASY repository,\footnote{\url{https://easy.dans.knaw.nl}} from ten archaeological projects \citep{ENKN-09,VENO13-08,MONF-09,VEEE-07,GOUA-08,VENO-02,KATK-08,WIJD-07,OOST-10,VEGL-10}, a total of 56583 geometries was collected in ten classes. Of the three tasks, the class distribution for this dataset is the most unbalanced: the data shows a clear over-representation of post holes (44,2 \%). Although the archaeological data task was designed for our experimental model evaluation purposes primarily, the ability to infer feature types from archaeological feature shapes is useful in field work, where a good performing model can assist in a documentation system, suggesting a likely feature class based on shape information.

\section{Evaluation}
\label{sec:results}
The accuracy scores of the model performance on each of the three tasks allow us to compare the performance of the deep learning models against the shallow learning models. Table~\ref{tab:evaluation} shows the results for each of the three benchmark tasks. The accuracy scores were produced from model predictions on the test set, consisting of geometries in the data set that were unseen by the models during training. The deep learning model experiments were repeated ten-fold: randomized network initialisation and batch sampling produce slight variations in accuracy scores between training sessions. The accuracy figures for the deep neural models therefore represent mean and standard deviation from the test predictions on the independently repeated training sessions.  

\begin{table}
\begin{tabular}{l p{0.2\textwidth} p{0.2\textwidth} p{0.2\textwidth}}
 Method & \multicolumn{3}{ l }{Task (no. of classes) } \\ \toprule
 & Neighbourhood \linebreak inhabitants (2) & Building types (9) & Archaeological \linebreak feature types (10) \\ \midrule
 Majority class		& 0.514 & 0.142 & 0.444 \\ \midrule
 k-NN 				& 0.671 & 0.377 & 0.596 \\
 Logistic regression& 0.659 & 0.328 & 0.555 \\
% SVM linear 		& 0.642 & 0.311 & 0.574 \\
% SVM polynomial 	& 0.631 & 0.317 & 0.574 \\
 SVM RBF 			& $\mathbf{0.683}$ & 0.365 & 0.601 \\ 
 Decision tree 		& 0.682 & 0.389 & 0.615 \\ \bottomrule
% CNN (fixed)		& $\mathbf{0.675 \pm 0.017}$ & $\mathbf{0.384 \pm 0.005}$  & $\mathbf{0.595 \pm 0.008}$ \\ 
 CNN				& $0.664 \pm 0.005$ & $\mathbf{0.408 \pm 0.003}$  & $\mathbf{0.624 \pm 0.002}$ \\ 
% RNN (fixed)		& $\mathbf{0.677 \pm 0.015}$ & $\mathbf{0.411 \pm 0.006}$  & $\mathbf{0.629 \pm 0.008}$ \\ 
 RNN				& $0.608 \pm 0.016$ & $0.389 \pm 0.008$  & $0.614 \pm 0.004$ \\ 
\end{tabular}
\caption{Table of results with accuracy scores for the majority class (top row) shallow models (middle four rows) the deep learning models (bottom two rows), with the best scores per task in bold. The number of classes per task is listed between brackets in the column headers. The standard deviations on the deep learning models on the bottom two rows were obtained from test set predictions on ten-fold repeated, independent training sessions.}
\label{tab:evaluation}
\end{table}

We note the following conclusions from these accuracy scores:
\begin{enumerate}
\item The deep neural nets are at least competitive with the best shallow models, for each of the three tasks. In five out of the six deep learning experiments, the deep models perform on par with or slightly better than the best shallow models, but in the broad sense they do not outperform the shallow models by a wide margin. 
\item On two of the three classification tasks, the CNN architecture is able to outperform the shallow models by a few percentage points. If top performance in a certain geometry classification task is required, the CNN is likely to be a good choice. 
\end{enumerate}

To gain further insight into model performance, we include an analysis the confusion matrices of the test runs. As there are 18 confusion matrices in total, these are not included in the article.\footnote{The confusion matrices are available for download from \url{https://dataverse.nl/api/access/datafile/13051}} In general the misclassification is reflected in similar patterns across all models, with higher errors for models that under-perform on a certain task. A few task-specific details are discussed in \ref{sec:confusion}, but in general the misclassification behaviour of the deep models  does not differ from the shallow models.

As mentioned in Section~\ref{sec:preprocessing}, the number of elliptic Fourier descriptors used for training the shallow models was included as a hyperparameter in the grid search. A closer inspection of these hyperparameters in Appendix~\ref{sec:baseline_hyperparams} is of interest:
\begin{enumerate}
\item Nearly all shallow models benefit from adding Fourier descriptors. A notable exception is the k-nearest neighbours algorithm, which scores the highest accuracy in two of the three tasks only when no Fourier descriptors are added to the training data. As it appears from the three tasks, the k-NN algorithm is less able to extract meaningful information from the Fourier descriptors.
\item The shallow models have a clear preference for lower orders of Fourier descriptors. Even though many higher orders (up to order 24) were tested, no shallow model was able to perform better on descriptor orders higher than four. Order four descriptors only provide a very rough approximation of the original geometry, as is well visualised in Kuhl and Giardina's paper \citet[243]{kuhl1982elliptic} and Figure~\ref{fig:efd_orders}. Still, the descriptors evidently contain enough important shape information for most shallow models to improve the accuracy score.
\item Support vector machines come with a misclassification tolerance hyperparameter $C$. In situations where SVMs with high $C$ settings (low tolerance) lead to higher performance, such as the ones for the archaeology classification task, the training sessions were exceedingly time-consuming to train on our data. Where low $C$-values tended to converge in seconds, high values could literally take days or even weeks to converge. To prevent having to wait for extended periods of time---there is no indication in what time frame a training session on a set of hyperparameters will converge---we needed to constrain the amount of training data and the maximum of iterations, especially on hyperparameter grid searches. It is quite possible that as a consequence of these constraints, the grid search fails to produce the optimal hyperparameter settings, but this is an unfortunate side effect of using SVMs on Fourier descriptors of geometries. 
\end{enumerate}

\section{Conclusion and future research}
\label{sec:conclusion}
In this article, we compared the accuracy of deep learning models against baselines of shallow learning methods on geospatial vector data classification tasks. We evaluated two deep learning models and four shallow learning models on three new classification tasks involving only geometries. To answer the question whether deep learning performs better than shallow learning, we directly compared the accuracy scores of these models on these three tasks. From our experiments we showed that our deep learning models are competitive with established shallow machine learning models, in two of the three tasks outperforming them by a small margin. 

None of the chosen recognition tasks appear to be trivial. Classifying objects from geometries alone is a tough assignment for any algorithm. The advantage of having a set of tough tasks is that these can serve as benchmarks: in future experiments, different learning algorithms or model layouts it may be possible to obtain higher accuracy scores. However, there is a possibility that the accuracy figures presented in the evaluation represent the maximum that can be learned from geometries alone. From these experiments alone it cannot be deduced whether these figures can actually be improved on. If there is a hard ceiling at the best performing models, perhaps the benchmarks can be improved by including more data than just the geometries alone, for example information gathered from the direct spatial surroundings or other properties of the spatial objects. The benchmark presented here can be considered a first attempt.

An area that might see improvement is the performance of LSTMs. In an earlier development stage, the LSTMs were trained on fixed length rather than on the variable length sequences. During this stage, the LSTMs performed significantly better (on validation data, no final tests were performed) on fixed length sequences, outperforming the CNNs. However, training on fixed length sequences was abandoned because it requires simplifying geometries to a fixed maximum of points per geometry. This was not consistent with our aim of training on all available information. Creating fixed length sequences also required adding a large amount of zero-padding to increase sequence length on all geometries shorter than the fixed size. After switching to variable length sequences, the performance of the CNN models increased and the LSTM performance dropped considerably. We hypothesize that there is room to improve the LSTM model configuration to CNN model performance or perhaps even better. To test this in future research, the fixed length sequences were included in the benchmark data. 

There are several open questions to further explore the use of deep learning models for geometries. It would be helpful to verify the accuracy on other types of geometries, such as multi-lines, multi-points or even heterogeneous geometry collections. Also, the deep neural net's comprehension of holes in polygons could be beneficial. Another interesting road to explore is to combine geometries with other information sources as input data, such as remote sensing data or textual descriptions. Deep learning poses a viable route to explore more complex pipelines. Such pipelines could include geometries and other modalities as inputs, to produce multi-modal combinations of sequences \citep{sutskever2014sequence}, images \citep{DBLP:journals/corr/HeGDG17} or new geometries \citep{ha2018a} as output. This paper is a step in that direction, by showing that is possible to have deep neural nets learn from geometries directly. 

\begin{appendices}
\section{Hyperparameter grid search results for shallow models}
\label{sec:baseline_hyperparams}

\begin{table}[!htb]
\begin{tabular}{p{0.15\textwidth} p{0.22\textwidth} p{0.22\textwidth} p{0.22\textwidth}}
 Method & \multicolumn{3}{ l }{Task} \\ \toprule
 & Neighbourhood \linebreak inhabitants (2) & Building types (9) & Archaeological \linebreak feature types (10) \\ \midrule
  \multirow{2}{*}{Decision tree} & $o$=2 & $o$=3 & $o$=3 \\
  & $d$=6 in $\left[4, 9\right]$ & $d$=10 in $\left[6, 12\right]$ & $d$=9 in $\left[5, 10\right]$ \\
 \hline
 \multirow{3}{*}{k-NN} & $o$=1 & $o$=0 & $o$=0 \\
 & k=26 in $\left[21, 30\right]$ & k=29 in $\left[21, 30\right]$ & k=29 in $\left[21, 30\right]$ \\
 \hline
 \multirow{3}{*}{SVM RBF} & o=1 & $o$=0 & $o$=2 \\
 & $C$=1 in 1e$\left[-2, 3\right]$ & C=1000 in 1e$\left[-2, 3\right]$ & C=100 in 1e$\left[-1, 3\right]$ \\
 & $\gamma$=1 in 1e$\left[-3, 3\right]$ & $\gamma$=10 in 1e$\left[-2, 3\right]$ & $\gamma$=0.01 in 1e$\left[-4, 4\right]$\\ 
 \hline
% SVM polynomial 		& $o$=1, $C=1000, degree=2	& $o$=1, $C$=1000, degree=1 & $o$=2, $C$=1000, degree=2 \\
% \hline
% SVM linear 			& $o$=1, $C$=1				& $o$=1, $C$=100			& $o$=4, C=100 \\
% \hline
 \multirow{2}{*}{\begin{tabular}[x]{@{}l@{}}Logistic\\regression\end{tabular}} & $o$=1 & $o$=4 & $o$=8 \\
 & $C$=0.01 in 1e$\left[-3, 1\right]$ & $C$=1 in 1e$\left[-2, 3\right]$ & $C$=1000 in 1e$\left[-2, 3\right]$ \\

\end{tabular}
\caption{Hyperparameters for the shallow models. Interval values for decision tree and k-nearest neighbours $\in \mathbb{N}$, for the SVM in log scale, with the exponent interval $\in \mathbb{N}$.}
\label{tab:baseline_hyperparameters}
\end{table}

The grid searches discussed in Section~\ref{sec:methods} resulted in a set of best hyperparameter settings for the shallow models. These best settings are listed in  Table~\ref{tab:baseline_hyperparameters} and include the ranges that were searched. The range for the elliptic fourier descriptor order $o$ is always the same: each grid search was executed on the orders $\langle0, 1, 2, 3, 4, 6, 8, 12, 16, 20, 24\rangle$. The search intervals for the other hyperparameters are listed in Table~\ref{tab:baseline_hyperparameters}. The k-hyperparameter of the k-nearest neighbour models and the maximum depth $d$ hyperparameter for the decision tree have intervals with values $\in \mathbb{N}$. For the other hyperparameters, the listed interval values are powers of ten, as indicated by the scientific notation. 

\section{Confusion matrix discussion}
\label{sec:confusion}
There are a few task-specific details to the confusion matrices mentioned in Section~\ref{sec:results}. Interesting to note is that all models over-estimate the number of inhabitants on the test set. Due to random selection, there is a slight over-representation of neighbourhoods below the median in the test set (51,4 \%) but this does not account for the large over-estimation of the models. Over-estimations are about twice (1,92) as frequent as the number of under-estimated errors. We can only speculate on why this is---it may be due to some shape imbalance in the random test set selection. All models struggle to distinguish between buildings with lodging function and habitation, which is understandable since any house can be made suitable for temporary lodging. Interestingly, the RNN model appears less prone to identifying lodging as habitation, but only to lose this advantage to a higher misclassification of habitation as lodging. Similarly, on the archaeology task all models have particular trouble separating natural phenomena from post holes, which can be attributed to being of similar size and shape. Often many archaeological features are preliminary marked as post holes when later, after making cross-sections, to be identified as natural phenomena. 

\end{appendices}

\bibliography{main}
\end{document}